\documentclass[final]{cvpr}

\usepackage{times}
\usepackage{epsfig}
\usepackage{graphicx}
\usepackage{amsmath}
\usepackage{amssymb}
\usepackage{array}
\usepackage{caption}
\usepackage{xcolor}
\usepackage[font=small,labelfont=bf,tableposition=top]{caption}
\usepackage{booktabs}
\usepackage{multirow}
\usepackage{xspace}
\usepackage{subfigure}
\usepackage{cite}
\usepackage{booktabs}
\usepackage{threeparttable}
\usepackage{microtype}
\usepackage{soul}

\def\etal{\emph{et al.}}
\definecolor{graycolor}{gray}{0.95}

\usepackage[pagebackref=false,breaklinks=true,colorlinks,bookmarks=false]{hyperref}

\def\cvprPaperID{265} 
\def\confYear{CVPR 2021}

\begin{document}

\title{Repetitive Activity Counting by Sight and Sound}

\author{Yunhua Zhang$^{1}$~~~~Ling Shao$^{2}$~~~~Cees G. M. Snoek$^{1}$\\[1mm]
\normalsize $^{1}$University of Amsterdam~~~$^{2}$Inception Institute of Artificial Intelligence
}



\maketitle

\begin{abstract}
This paper strives for repetitive activity counting in videos. Different from existing works, which all analyze the visual video content only, we incorporate for the first time the corresponding sound into the repetition counting process. This benefits accuracy in challenging vision conditions such as occlusion, dramatic camera view changes, low resolution, etc. We propose a model that starts with analyzing the sight and sound streams separately. Then an audiovisual temporal stride decision module and a reliability estimation module are introduced to exploit cross-modal temporal interaction. For learning and evaluation, an existing dataset is repurposed and reorganized to allow for repetition counting with sight and sound. We also introduce a variant of this dataset for repetition counting under challenging vision conditions. Experiments demonstrate the benefit of sound, as well as the other introduced modules, for repetition counting. Our sight-only model already outperforms the state-of-the-art by itself, when we add sound, results improve notably, especially under harsh vision conditions. The code and datasets are available at \url{https://github.com/xiaobai1217/RepetitionCounting}. 
\end{abstract}

\begin{figure}[t!]
\centering
\includegraphics[width=0.99\linewidth]{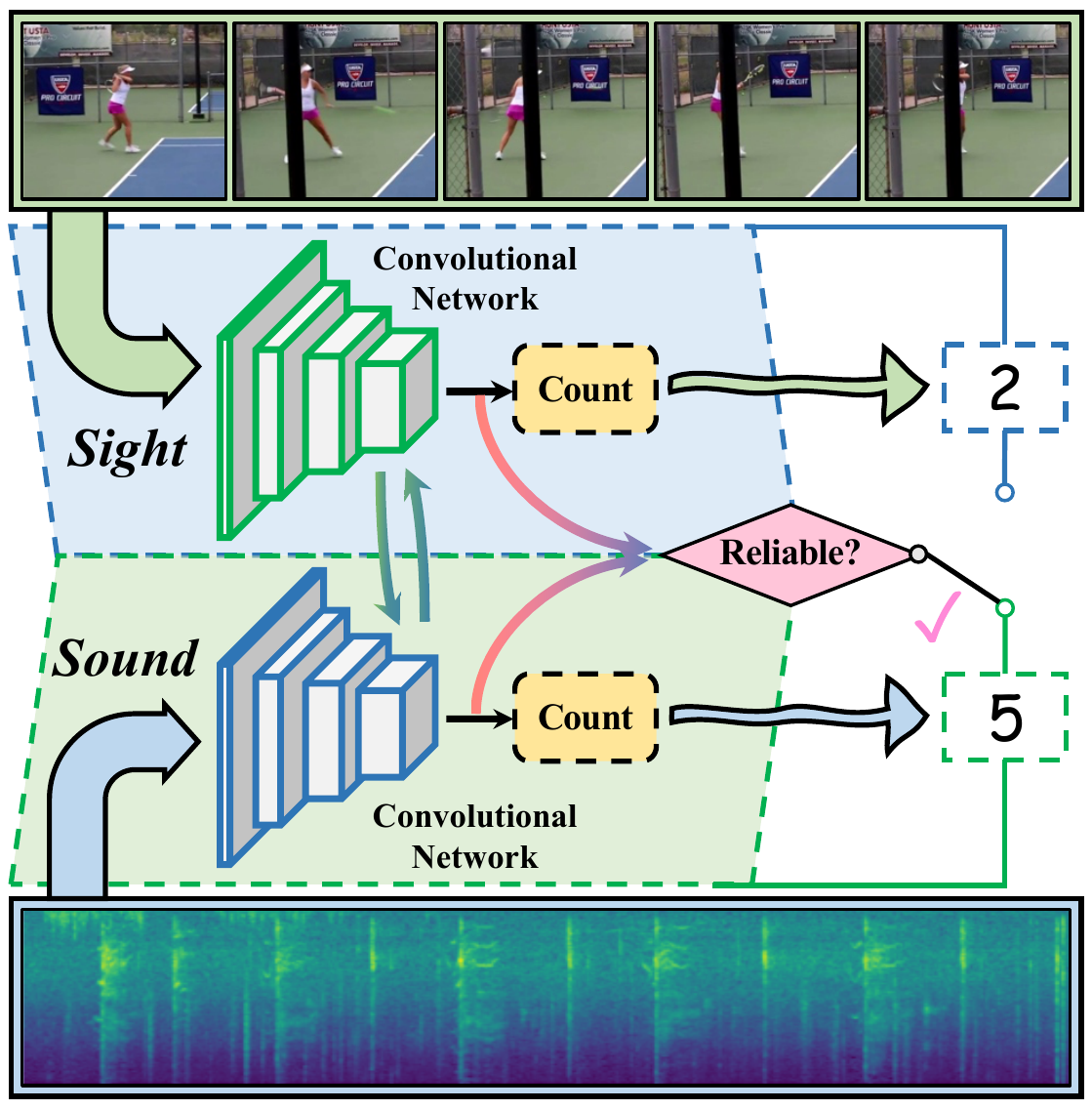}
\caption{From sight and sound, as well as their cross-modal interaction, we predict the number of repetitions for an (unknown) activity happening in a video. This is especially beneficial in challenging vision conditions with occlusions and low illumination.}
\label{fig:1stFigure}
\end{figure}

\section{Introduction}
The goal of this paper is to count in video the repetitions of (unknown) activities, like bouncing on a trampoline, slicing an onion or playing ping pong. The computer vision solutions to this challenging problem have a long tradition. Early work emphasized on repetitive motion estimation by Fourier analysis, \eg,~\cite{arnold2008,fourier1,fourier2,fourier3}, and more recently by a continuous wavelet transform~\cite{Tom2018,TomIJCV}. State-of-the-art solutions rely on convolutional neural networks~\cite{Levy2015,context2020, zisserman2020} and large-scale count-annotated datasets~\cite{context2020, zisserman2020} to learn to predict the number of repetitions in a video. Albeit successful, all existing works focus exclusively on the visual modality, and could fail in poor sight conditions such as low illumination, occlusion, camera view changes, \etc. Different from existing works, we propose in this paper the first repetitive activity counting method based on sight \textit{and} sound.

Analyzing sound has recently proven advantageous in a variety of computer vision challenges, such as representation learning by audio-visual synchronization~\cite{korbar2018cooperative, arandjelovic2017look, aytar2016soundnet,owens2016ambient}, video captioning~\cite{wang2018watch,tian2018attempt,rahman2019watch}, sound source localization~\cite{senocak2019learning,rouditchenko2019self}, to name a few. Correspondingly, several mechanisms for fusing both modalities have been introduced. In works for action recognition by previewing the audio track~\cite{korbar2019scsampler} and talking-face generation~\cite{kaisiyuan2020mead,song2020everybody}, the audio network usually works independently and the predictions guide the inference process of the visual counterpart.  In contrast, feature multiplication and concatenation operations, as well as cross-modal attention mechanisms, are widely adopted for fusion in tasks like audio-visual synchronization~\cite{korbar2018cooperative, rouditchenko2019self, arandjelovic2018objects} and video captioning~\cite{wang2018watch,tian2018attempt,rahman2019watch}. We also combine sight and sound, but observe that for some activities, like playing ping pong, humans can count the number of repetitions by just listening. This gives us an incentive that sound could be an important cue by itself. Hence, an intelligent repetition counting system should be able to judge when the sight condition is poor and therefore utilize complementary information from sound.

The first and foremost contribution of this paper is addressing video repetition estimation from a new perspective based on not only the sight but also the sound signals. 
As a second contribution, we propose an audiovisual model with a sight and a sound stream, where each stream facilitates each modality to predict the number of repetitions. As the repetition cycle lengths may vary in different videos, we further propose a temporal stride decision module to select the best sample rate for each video based on both visual and audio features. Our reliability estimation module finally exploits cross-modal temporal interaction to decide which modality-specific prediction is more reliable. Since existing works focus on visual repetition counting only, our third contribution entails two sight and sound datasets that we derive from Countix~\cite{zisserman2020} and VGGsound~\cite{VGGSound}. 
One of our datasets is for supervised learning and evaluation and the other for assessing audiovisual counting in various challenging vision conditions. Finally, our experiments demonstrate the benefit of sound, as well as the other introduced network modules, for repetition counting. Our sight-only model already outperforms the state-of-the-art by itself, and when we add sound, the results improve further, especially under harsh vision conditions. Before detailing our model, as summarized in Figure~\ref{fig:1stFigure}, we first discuss related work.

\section{Related Work}
\noindent \textbf{Repetitive activity counting.} 
Existing approaches for repetition estimation in video rely on visual content only. Early works~\cite{arnold2008,fourier1,fourier2, fourier3} compress the motion field of video into one-dimensional signals and count repetitive activities by Fourier analysis~\cite{arnold2008,fourier1,fourier2, fourier3}, peak detection~\cite{thangali2005periodic} or singular value decomposition~\cite{chetverikov2006motion}. Burghouts and Geusebroek~\cite{burghouts2006quasi} propose a spatiotemporal filter bank, which works online but needs manual adjustment. Levy and Wolf~\cite{Levy2015} design a classification network able to learn from synthetic data. Their network is designed to extract features from an input video with a predefined sampling-rate, which cannot handle repetitions with various period lengths. The synthetic dataset is also less suitable for usage in the wild. All of the above methods assume the repetitions are periodic, so they can cope with stationary situations only. 

Recently, algorithms for non-stationary repetitive action counting have been proposed. Runia \etal~\cite{Tom2018, TomIJCV} are the first to address non-stationary situations. They leverage the wavelet transform based on the flow field and collect a dataset containing $100$ videos including non-stationary repetitions, but the videos do not contain an audio track. 
Zhang \etal~\cite{context2020} propose a context-aware framework based on a 3D convolution network, and introduce a new activity repetition counting dataset based on UCF101~\cite{soomro2012ucf101}. While effective, the temporal length of every two repetitions is predicted by an iterative refinement, making the approach less appealing from a computational perspective. Dwibedi \etal~\cite{zisserman2020} collect a large-scale dataset from YouTube, named Countix, containing more than 6,000 videos with activity repetition counts. Their method utilizes temporal self-similarity between video frames for repetition estimation. It chooses the frame rate sampling the input video by picking the one with the maximum periodicity classification score. While appealing, such a rate selection scheme is not optimal for accurate counting, as it is prone to select high frame rates leading to omissions. 

Different from all these existing methods, we propose to address repetitive activity counting by sight and sound. Our network contains a temporal stride decision module able to choose the most suitable frame rate for counting, based on features from both modalities. To facilitate our investigation, we reorganize and supplement the Countix~\cite{zisserman2020} dataset, to arrive at two audiovisual datasets for repetitive activity counting by sight and sound.

\noindent \textbf{Learning by sight and sound.} 
Many have demonstrated the benefit of audio signals for various computer vision challenges, \eg, action recognition~\cite{korbar2019scsampler,gao2020listen}, audiovisual event localization~\cite{wu2019dual} and self-supervised learning~\cite{korbar2018cooperative, arandjelovic2017look, aytar2016soundnet,owens2016ambient}. 
As processing audio signals is much faster than video frames, both Korbar \etal~\cite{korbar2019scsampler} and Gao \etal~\cite{gao2020listen} reduce the computational cost by previewing the audio track for video analysis. 
However, the Kinetics-Sound~\cite{arandjelovic2017look} used for training and evaluation in~\cite{arandjelovic2017look, korbar2019scsampler, gao2020listen} is simply formed by all the videos in the Kinetics dataset~\cite{Kinetics} covering 34 human action classes, which are potentially manifested visually and aurally. As a result, the audio track of many videos is full of background music, which introduces noise for training and fair evaluation. 
Recent talking-face generation works exploit sound for creating photo-realistic videos~\cite{kaisiyuan2020mead,song2020everybody}. While audio features are used to generate expression parameters in~\cite{song2020everybody}, Wang~\etal~\cite{kaisiyuan2020mead} to map the audio to lip movements. 
There are also numerous works~\cite{joze2020mmtm,tsiami2020stavis,korbar2018cooperative,cartas2019seeing,senocak2019learning,xu2019recursive} that consider the interaction between both modalities. Some simply integrate features by concatenation for tasks like saliency detection~\cite{tsiami2020stavis} and self-supervised learning~\cite{korbar2018cooperative, rouditchenko2019self, arandjelovic2018objects}. %
Cartas \etal~\cite{cartas2019seeing} combine multi-modal predictions by averaging or training a fully connected layer independently, for egocentric action recognition. 
Works for sound source localization~\cite{senocak2019learning,rouditchenko2019self} and separation~\cite{xu2019recursive,gao2019co,gao2018learning,thesoundofmotion} also commonly generate cross-modal attention maps. 
Hu \etal~\cite{audiovisualcrowd} use audio features to modulate the visual features for more accurate crowd counting. 

Exploiting sound for activity repetition counting is still unexplored and to obtain audiovisual datasets facilitating our research, we also select videos with usable audio track from a large scale visual-only dataset manually to reduce label noise. To cope with various `in the wild' conditions, we further introduce a novel scheme to explicitly estimate the reliability of the predictions from sight and sound.

\begin{figure*}[t!]
\centering
\includegraphics[width=1.0\linewidth,height=0.4\linewidth]{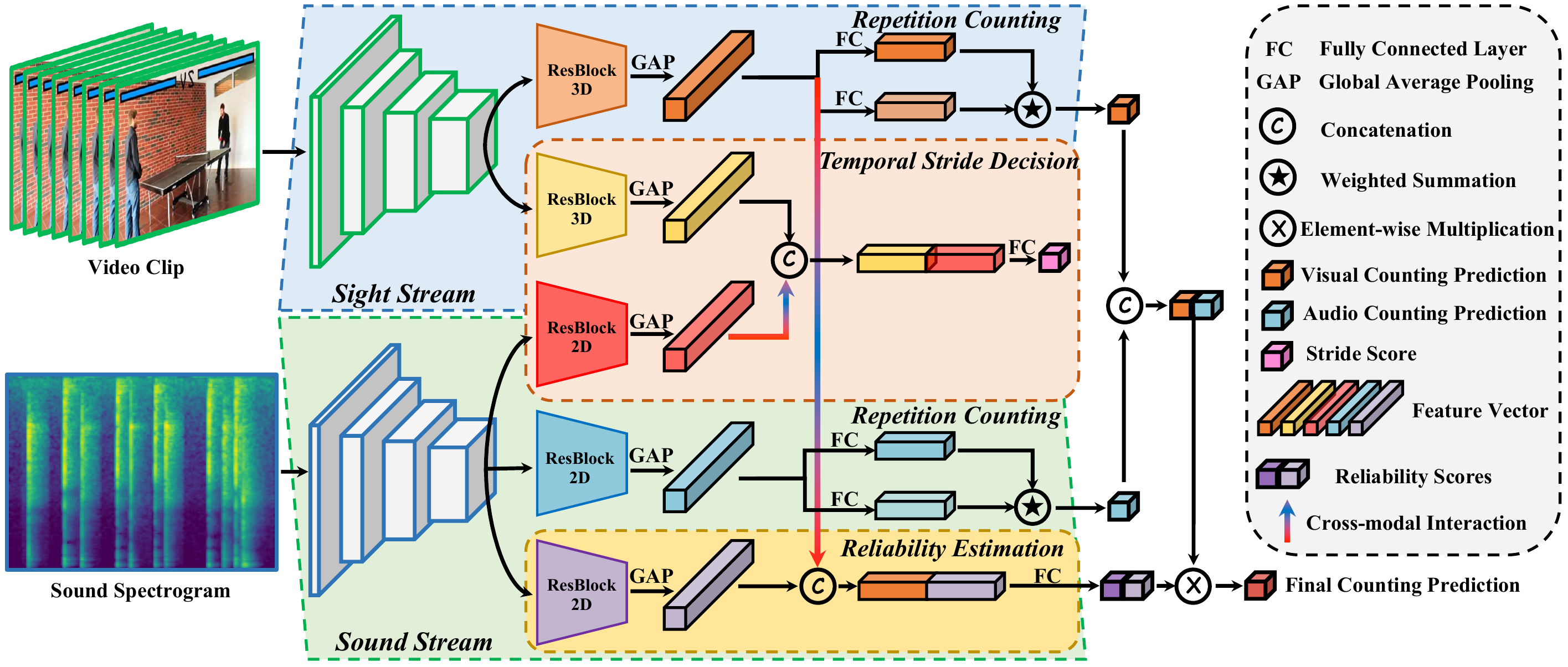}
\caption{\textbf{Proposed class-agnostic activity repetition counting model.} Our model contains four components (1) sight stream, (2) sound stream, (3) temporal stride decision module and (4) reliability estimation module. Both streams contain a backbone network that outputs a modality-specific counting prediction. The temporal stride decision module takes audio and visual features as inputs and outputs the frame sample rate for the input video. Finally, the reliability estimation module decides what prediction from which modality to use.}
\label{fig:framework}
\end{figure*}

\section{Model}
Given a video, containing a visual stream and its corresponding audio stream, our goal is to count the number of repetitions of (unknown) activities happening in the content. To achieve this, we propose a model that contains four modules. (\textit{i}) The sight stream adopts a 3D convolutional network as the backbone. It takes video clips as inputs and outputs the counting result for each clip. (\textit{ii}) For the sound stream, we rely on a 2D convolutional network, which takes the sound spectrogram generated by the short-time Fourier transform as input and outputs the counting result in the same way as the sight stream. (\textit{iii}) A temporal stride decision module is designed to select the best temporal stride per video for the sight stream based on both visual and audio features. (\textit{iv}) Finally, the reliability estimation module decides what prediction to use. The overall model is summarized in Figure~\ref{fig:framework} and detailed per module next. 

\subsection{Repetition Counting by Sight}
The sight stream uses an S3D~\cite{S3D} architecture and the final classification layer is replaced by two separate fully connected layers, as shown in Figure~\ref{fig:framework}. 
Given a video clip $V_i$ of size $T{\times} H {\times} W {\times} 3$, visual features are extracted with the following equation:
\begin{equation}
\mathbf{v}_{i,\textit{feat}} = \mathcal{V}_{\textit{CNN}}(V_i),
\end{equation}
where $\mathbf{v}_{i,\textit{feat}} \in \mathbb{R}^{512}$. 
Intuitively, a fully connected layer with one output unit could suffice to output the counting result. However, this setting leads to inferior repetition counts since different types of movements should not be counted in the same way, and each action class cannot be simply regarded as one unique repetition class. For example, different videos of doing aerobics contain various repetitive motions, while bouncing on a bouncy castle or a trampoline contains similar movements despite belonging to different action classes. 
Therefore, in our work, two fully connected layers work in tandem, with one $f_v^1$ outputting the counting result of each repetition class and the other one $f_v^2$ classifying which repetition class the input belongs to: 
\begin{equation}
\begin{split}
& \mathbf{C}_{i,v}^{'} = f_v^1(\mathbf{v}_{i,\textit{feat}}), \mathbf{C}_{i,v}^{'} \in \mathbb{R}^P, \\
& \mathbf{T}_{i,v} = \textit{softmax}(f_v^2(\mathbf{v}_{i,\textit{feat}})), \mathbf{T}_{i,v} \in \mathbb{R}^P,
\label{eq:count1}
\end{split}
\end{equation}
where $P$ is the number of repetition classes, $\mathbf{C}_{i,v}^{'}$ is the counting result of each class, and $\mathbf{T}_{i,v}$ is the classification result by the softmax operation. Here, we assume that there are roughly $P$ classes of repetitive motion patterns, and the network learns to classify the training videos into those $P$ classes automatically during training. 
Then the final counting result $\mathbf{C}_{i,v}$ from the visual content is obtained by:
\begin{equation}
C_{i,v} = \sum_{k=1}^{P}\mathbf{C}_{i,v}^{'}(k)\mathbf{T}_{i,v}(k). 
\label{eq:count2}
\end{equation}
For training the repetition counting, we define the loss function as follows:
\begin{equation}
L^{'} = \frac{1}{N}\sum_{i=1}^N L_2(C_{i,v}, l_{i}) + \lambda_1^v \frac{|C_{i,v}-l_{i}|}{l_{i}},
\end{equation}
where $N$ is the batch size, $L_2$ is the L2 loss~\cite{rothe2018deep}, $l_{i}$ is the groundtruth count label of the $i$th sample and $\lambda_1^v$ is a hyperparameter for the second term. Note that when using the L2 loss only, the model tends to predict samples with groundtruth counts of large values accurately, due to higher losses, while for videos with a few repetitions, the predicted counts tend to be unreliable. Therefore, we add a second term here to let the model pay more attention to such data. 

Besides, we expect the output units of $f_v^2$ to focus on different repetition classes given various videos. However, without constraint, $f_v^2$ could simply output a high response via the same unit. 
To avoid such degenerated cases, 
we add a diversity loss~\cite{cosineloss} based on the cosine similarity: 
\begin{equation}
L_{i,v}^{\textit{div}} = \sum_{q=1}^{P-1}\sum_{j=q+1}^{P}\frac{T_{i,v}^q\cdot T_{i,v}^j}{||T_{i,v}^q||||T_{i,v}^j||}, 
\end{equation}
where $T_v^q$ and $T_v^j$ are the $q$th and $j$th units of the classification outputs. 
By minimizing such a diversity
loss, the output $T_v$ in the same batch are encouraged to produce different activations on different types of repetitive motions. 
Then the total loss function is: 
\begin{equation}
\label{eq:loss_final_v}
L_v = \frac{1}{N}\sum_{i=1}^N L_2(C_{i,v}, l_{i}) + \lambda_1^v \frac{|C_{i,v}-l_{i}|}{l_{i}} + \lambda_2^v L_{i,v}^{\textit{div}}, 
\end{equation}
where $\lambda_2^v$ is a hyperparameter.

\subsection{Repetition Counting by Sound}
The sound stream adopts a ResNet-18~\cite{resnet} as the backbone. Following~\cite{audioclassification, VGGSound}, we first transform the raw audio clip into a spectrogram and then divide it into a series of $257 {\times} 500$ spectrograms, which become the inputs to our network. Similar to the sight stream, we also replace the final classification layer by two separate fully connected layers, with one classifying the input and the other one outputting the corresponding counting result of each repetition class. We use the same loss function as the sight stream:
\begin{equation}
\label{eq:loss_final_a}
L_a=\frac{1}{N}\sum_{i=1}^N L_2(C_{i,a}, l_{i}) + \lambda_1^a \frac{|C_{i,a}-l_{i}|}{l_{i}} + \lambda_2^a L_{i,a}^{\textit{div}}, 
\end{equation}
where $C_{i,a}$ is the counting result from the audio track, and $\lambda_1^a$ and $\lambda_2^a$ are hyperparameters.

\subsection{Temporal Stride Decision} \label{sec:stride}
Repetitions have various period lengths for different videos. For the sound stream, we can simply resize the spectrogram along the time dimension to ensure each $257 {\times} 500$ segment to have at least two repetitions. 
However, for the sight stream, we cannot roughly resize the video frames along the time dimension. Therefore, for each video, we need to use a specific temporal stride (\ie frame rate) to form video clips of $T$ frames as the inputs. This is important as video clips with small temporal strides may fail to include at least two repetitions, while too large temporal strides lead the network to ignore some repetitions. Therefore, we add an additional temporal stride decision module to select the best temporal stride for each video. It has two parallel residual blocks, processing visual and audio features from the third residual block of the two streams, with the same structure as those of the backbones. 
Then we concat the output features (as shown in Figure~\ref{fig:framework}) and send them into a fully connected layer, which outputs a single unit representing the score of the current temporal stride. We use a max-margin ranking loss for training this module: 
\begin{equation}
L_s = \frac{1}{N}\sum_{i=1}^N \textit{max}(0, s_i^- - s_i^+ +m), 
\label{eq:m}
\end{equation}
where $m$ is the margin, $s_i^-$ and $s_i^+$ are the scores from negative and positive strides. During inference, we send a series of clips from the same video with different strides into the network, and select the stride with the highest score. 

\noindent \textbf{Training details.} For each training video, the trained visual model predicts the counting result with a series of temporal strides, \ie $s={1,...,S_k,...,S_K}$, where $S_K$ is the maximum stride we use. 
Then we can obtain corresponding predictions ${C_{i,v}^1,...,C_{i,v}^{S_K}}$. First, we select the temporal strides that cover less than two repetitions as negative strides. Then, we choose the smallest stride that is enough to contain at least two repetitions as the positive temporal stride $S^*$. Correspondingly, 
for the remaining strides, we quantitatively compute their prediction deviations from the prediction of the positive stride by: 
\begin{equation}
\delta_n = \frac{C_{i,v}^{*}-C_{i,v}^k}{C_{i,v}^{*}},
\label{eq:stride}
\end{equation}
where $C_{i,v}^{*}$ and $C_{i,v}^{k}$ are the counting predictions from the best stride and a selected stride, 
and $\delta_n$ is the computed deviation. Finally, we select strides with $\delta_n{>}\theta_s$ ($\theta_s$ is a predefined threshold) as negative strides, since for these strides, the network begins to omit certain repetitions. 
During training, for each video, its $S^*$ is used to form a positive video clip outputting $s^+_i$, while we randomly select one from the negative strides to generate the clip outputting $s^-_i$. 

\subsection{Reliability Estimation} \label{sec:reliability}
Depending on the sensory video recording conditions, the reliability of the sight and sound predictions may vary. To compensate for this variability, we introduce a reliability estimation module to decide what prediction from which modality is more reliable for the current input. 
As shown in Figure~\ref{fig:framework}, it contains one residual block for processing the audio feature and one fully connected layer taking features from both modalities as inputs. The output is a single unit processed by a sigmoid function and represents the confidence $\gamma$ of the audio modality. 
Correspondingly, the confidence of the visual modality is $1-\gamma$. 
Then the final counting result is obtained by:
\begin{equation}
C_i = C_{i,v}*(1-\gamma) + C_{i,a}*\gamma. 
\label{eq:r_fusion}
\end{equation}
As $C_i$ is expected to be close to the groundtruth counting label, the loss function we use for training is: 
\begin{equation}
L_r = \frac{1}{N} \sum_{i=1}^N \frac{|C_i-l_i|}{l_i}. 
\label{eq:loss_r}
\end{equation}

\noindent \textbf{Training details.} 
During training, for each video, the accuracy of $C_{i,v}$ and $C_{i,a}$ in Eq.~\ref{eq:r_fusion} is expected to indicate the reliability of the corresponding modality content. 
To get $C_{i,v}$ and $C_{i,a}$, one simple approach is to directly use the predictions from the trained models. 
However, we empirically observe that such a manner suffers from severe over-fitting, %
since the learned models could overfit to recent training samples with poor modality content. As a result, the obtained $C_{i,v}$ and $C_{i,a}$ cannot represent the reliability effectively. 
However, for one modality of a video, if the corresponding model predicts inaccurately most of the time during training, then intuitively the content may be too noisy or poor for learning. 
Therefore, instead of $C_{i,v}$ and $C_{i,a}$ from the final models, we use the average prediction of each stream at different training stages. 
Here, we take the sight stream as an example. 
After each training epoch, if the loss computed by $\frac{1}{\mathcal{N}} \sum_{i=1}^{\mathcal{N}} \frac{|C_{i,v}-l_i|}{l_i}$, $\mathcal{N}$ is the number of videos, over the validation set is below a threshold $\theta_r^v$ (\ie current model parameters have competitive performance), we record the predictions of the model over the training videos. 
Once the training is finished, we can obtain the average prediction (\ie empirical prediction) of each training video by recordings correspondingly. 
The empirical prediction of the sound stream is computed in the same way, with a threshold $\theta_r^a$ for the validation loss. 
Finally, our reliability estimation module uses those empirical predictions for Eq.~\ref{eq:r_fusion} during training and learns to switch between sight and sound models for effective late fusion. 
%

\section{Experimental Setup}

\subsection{Datasets}
Existing datasets for repetition counting~\cite{Levy2015, Tom2018, context2020, zisserman2020} focus on counting by visual content only. Thus, the videos have either no audio information at all, or at best a few only. Nonetheless, we evaluate our (sight) model on the two largest existing visual-only datasets, \ie UCFRep and Countix. As we focus on counting by sight and sound, we also repurpose, reorganize and supplement one of those two datasets. 

\noindent \textbf{UCFRep.} The UCFRep dataset by Zhang \etal~\cite{context2020} contains 526 videos of 23 categories selected from UCF101~\cite{soomro2012ucf101}, a widely used benchmark for action recognition, with 420 and 106 videos for training and validation. Particularly, it has boundary annotations for each repetition along the time dimension. However, the large majority of videos do not have any associated audio track. 

\begin{table}[t!]
\centering
\begin{tabular}{lc}
\toprule
\textbf{Vision challenge} & \textbf{Number of videos} \\
\midrule
Camera viewpoint changes & 69 \\
Cluttered background & 36 \\
Low illumination & 13  \\
Fast motion & 31  \\
Disappearing activity & 25   \\
Scale variation & 24   \\
Low resolution & 29 \\
\midrule
\textit{Overall} & 214  \\
\bottomrule
\end{tabular}
\caption{\textbf{Extreme Countix-AV dataset statistics.}}
\label{tab:Extreme}
\end{table}

\noindent \textbf{Countix.} 
The Countix dataset by Dwibedi \etal~\cite{zisserman2020} serves as the largest dataset for video repetition counting in the wild. It is a subset of the Kinetics~\cite{Kinetics} dataset annotated with segments of repeated actions and corresponding counts. The dataset contains 8,757 videos in total of 45 categories, with 4,588, 1,450 and 2,719 for training, validation and testing. 

\noindent \textbf{Countix-AV.} 
We repurpose and reorganize the Countix dataset for our goal of counting repetitive activities by sight and sound. 
We first select 19 categories for which the repetitive action has a clear sound, such as \textit{clapping}, \textit{playing tennis}, etc. 
As several videos contain artificially added background music or have no audio track at all, they are less suited for assessing the impact of the sound-stream, and the sight and sound combination. Therefore, we manually filter out such videos so that the videos preserved are guaranteed to contain the environmental sound only, be it they may also include realistic background noise or unclear repetition sounds. 
This results in the Countix-AV dataset consisting of 1,863 videos, with 987, 311 and 565 for training, validation and testing. We maintain the original count annotations from Countix and keep the same split (\ie training, validation, or testing) for each video. The dataset is detailed in the appendix. %

\begin{figure*}[t!]
\centering
\includegraphics[width=1.0\linewidth,height=0.65\linewidth]{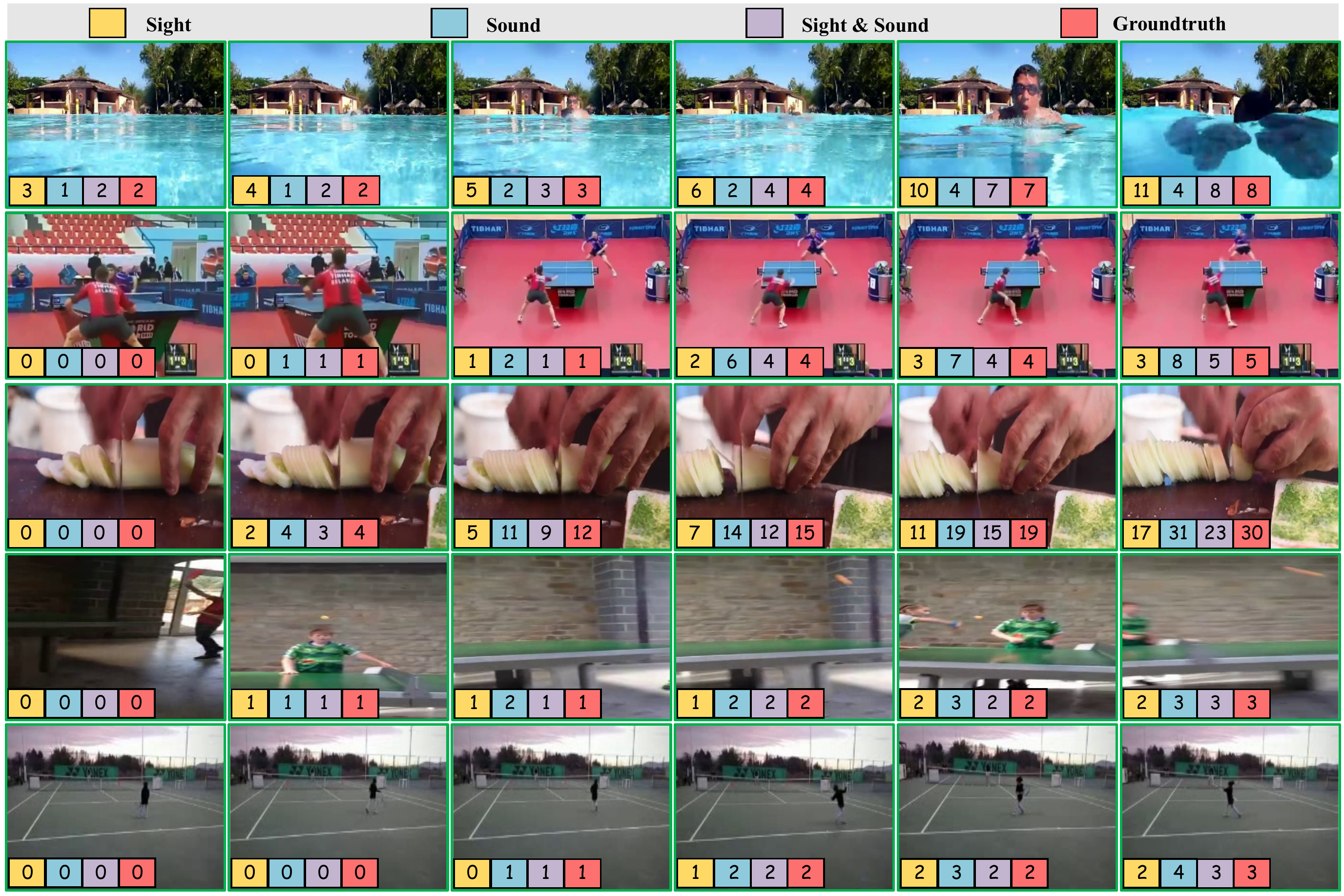}
\caption{\textbf{Example videos from the Extreme Countix-AV dataset.} From top-row to bottom-row are videos with scale variation, camera viewpoint change, fast motion, a disappearing activity and low resolution, with the numbers in colored boxes indicating counting results and the corresponding groundtruth. }
\label{fig:qualitative2}
\end{figure*}

\noindent \textbf{Extreme Countix-AV.} 
For most videos in Countix-AV, the ongoing action is both visible and audible. As the audio signal is expected to play a vital role when the visual content is not reliable, we further introduce the Extreme Countix-AV dataset to quantitatively evaluate the benefits brought by audiovisual counting under various extreme sight conditions. 
For data collection, we first define 7 vision challenges, according to which we select videos. Then, 156 videos from Countix-AV are selected and to enlarge this selection, we choose and label another 58 videos from the VGGSound dataset by Chen~\etal~\cite{VGGSound}. The overall dataset and challenges are summarized in Table~\ref{tab:Extreme}, with video examples depicted in Figure~\ref{fig:qualitative2}. More details are provided in the appendix.

\subsection{Evaluation Criteria}
\label{sec:evaluation_criteria}
We adopt the same evaluation metrics as previous works~\cite{arnold2008,Tom2018,TomIJCV,context2020,zisserman2020}, \ie the mean absolute error (MAE) and off-by-one accuracy (OBO), defined as follows:
\begin{equation}
    \textit{MAE} = \frac{1}{\mathcal{N}} \sum_{i=1}^{\mathcal{N}}\frac{|\hat{c_i} - l_i|}{l_i}, 
    \end{equation}
    \begin{equation}
    \textit{OBO} = \frac{1}{\mathcal{N}} \sum_{i=1}^{\mathcal{N}} [|\hat{c_i}-l_i|\leq 1],
\end{equation}
where $\mathcal{N}$ is the total number of videos, $\hat{c_i}$ is the model prediction of the $i$th video and $l_i$ is the groundtruth. 
Specifically, for the Extreme Countix-AV, we report MAE only, as those videos have more repetitions than other datasets and OBO cannot evaluate the performance effectively. 
\subsection{Implementation Details}
\label{sec:implementation_details}
We implement our method using PyTorch with two NVIDIA GTX1080Ti GPUs. 
We provide training details below and inference procedures in the appendix. 

\noindent \textbf{Sight and sound models.} For the sight stream, all input video frames are resized to $112{\times}112$, and we form each clip of $64$ frames with its temporal stride $S^*$ defined in Section~\ref{sec:stride}. We initialize the backbone with weights from a Kinetics~\cite{Kinetics} pre-trained checkpoint. The training of the sight model is on the original Countix training set~\cite{zisserman2020} and takes 8 epochs by SGD with a fixed learning rate of $10^{-4}$ and batch size of $8$. $\lambda^1_v$, $\lambda_v^2$, $\lambda_a^1$ and $\lambda_a^2$ are all set to $10$. The sound model is trained with the same setting as the sight stream but using our Countix-AV training set for 20 epochs. 

\noindent \textbf{Temporal stride decision module.} 
The training takes $5$ epochs with a learning rate of $10^{-3}$ after obtaining the negative strides of each video. Here, we provide two options. First, it can be trained with the visual modality only, \ie without the audio feature, using the original Countix~\cite{zisserman2020} dataset so that the sight model can work independently. The other option is our full setting (as shown in Figure~\ref{fig:framework}) trained on Countix-AV with the audio modality. Margin $m$ in Eq.~\ref{eq:m} is set to 2.9 and $S_K$ is set to 8.  In experiments, we find the value of $\theta_s$ does not influence results too much, and $\theta_s{=}0.29$ works best (see appendix for ablation). 

\noindent \textbf{Reliability estimation module.} 
We first collect the empirical predictions before training, and $\theta_r^v$ and $\theta_r^a$ are set to 0.36 and 0.40. Then, this module is trained on Countix-AV for 20 epochs with a learning rate of $10^{-4}$ and batch size of 8. 
%

\section{Results}

\noindent \textbf{Benefit of model components.} Our model consists of four main components: the sight and sound counting models, the temporal stride decision module and the reliability estimation module. We evaluate the performance of several network variants on Countix-AV to validate the efficacy of each component. The results are shown in Table~\ref{tab:ablation}. 
Note that for ``Sight stream" in the first row, its temporal stride decision module takes the visual modality only as input. 
In isolation, the sight stream performs better than the sound stream. When we incorporate audio features into the temporal stride decision module, denoted as ``Sight with temporal stride", the MAE of the sight stream is further reduced from 0.331 to 0.314. This demonstrates the audio signals provide useful temporal information. Simply averaging the predictions from both modalities results in higher accuracy than either modality alone. However, when we further reweigh the predictions by our reliability estimation module, we obtain the best result with an MAE of 0.291 and an OBO of 0.479. 

\begin{table}[t!]
\centering
\begin{tabular}{lcc}
\toprule
\textbf{Model components} & \textit{MAE~$\downarrow$} & \textit{OBO~$\uparrow$}  \\
\midrule
Sight stream & 0.331 & 0.431 \\
Sound stream & 0.375 & 0.377\\
Sight with temporal stride & 0.314 & 0.459 \\
Averaging predictions & 0.300 & 0.439 \\
Full sight and sound model & 0.291 & 0.479 \\
\bottomrule
\end{tabular}
\caption{\textbf{Benefit of model components} on Countix-AV. All modules matter and reliability estimation is preferred over simple averaging of sight and sound predictions.}
\label{tab:ablation}
\end{table}

\noindent \textbf{Influence of loss function terms.}
The loss function used for training the visual and audio models consists of three terms. We perform an ablation on different term combinations to further understand their contributions. Results in Table~\ref{tab:av_loss} indicate both $L_{\textit{div}}$ and $L_{\textit{mae}}$ reduce the counting error, especially on the sound stream. We observe adding $L_{\textit{div}}$ contributes to performance improvements because it allows the units in the classification layer to affect each other during training. It prevents this layer from converging to a degenerated solution, in which all videos are assigned to the same repetition class. Combining all loss terms during training produces best results for both modalities. 

\begin{table}[t!]
\resizebox{\linewidth}{!}{
\begin{tabular}{lcccc}
\toprule
 & \multicolumn{2}{c}{\textbf{Sight}} & \multicolumn{2}{c}{\textbf{Sound}}\\
 \cmidrule(lr){2-3} \cmidrule(lr){4-5}
\textbf{Loss term} & \textit{MAE~$\downarrow$}  & \textit{OBO~$\uparrow$} & \textit{MAE~$\downarrow$}  & \textit{OBO~$\uparrow$}\\
\midrule
$L_2$                   & 0.371 & 0.424 & 0.471 & 0.338 \\
$L_2+L_{\textit{div}}$           & 0.324 & 0.478 & 0.410 & 0.343 \\
$L_{\textit{div}}+L_{\textit{mae}}$       & 0.356 & 0.446 & 0.447 & 0.310 \\
$L_2+L_{\textit{mae}}$           & 0.370 & 0.421 & 0.426 & 0.340 \\
$L_2+L_{\textit{div}}+L_{\textit{mae}}$   & 0.314 & 0.498 & 0.375 & 0.377 \\
\bottomrule
\end{tabular}
}
\caption{\textbf{Influence of loss function terms} on Countix-AV. Each term lowers the counting error, while $L_{\textit{div}}$ is indispensable. }
\label{tab:av_loss}
\end{table}

\begin{figure}[b!]
\centering
\includegraphics[width=0.99\linewidth,height=0.8\linewidth]{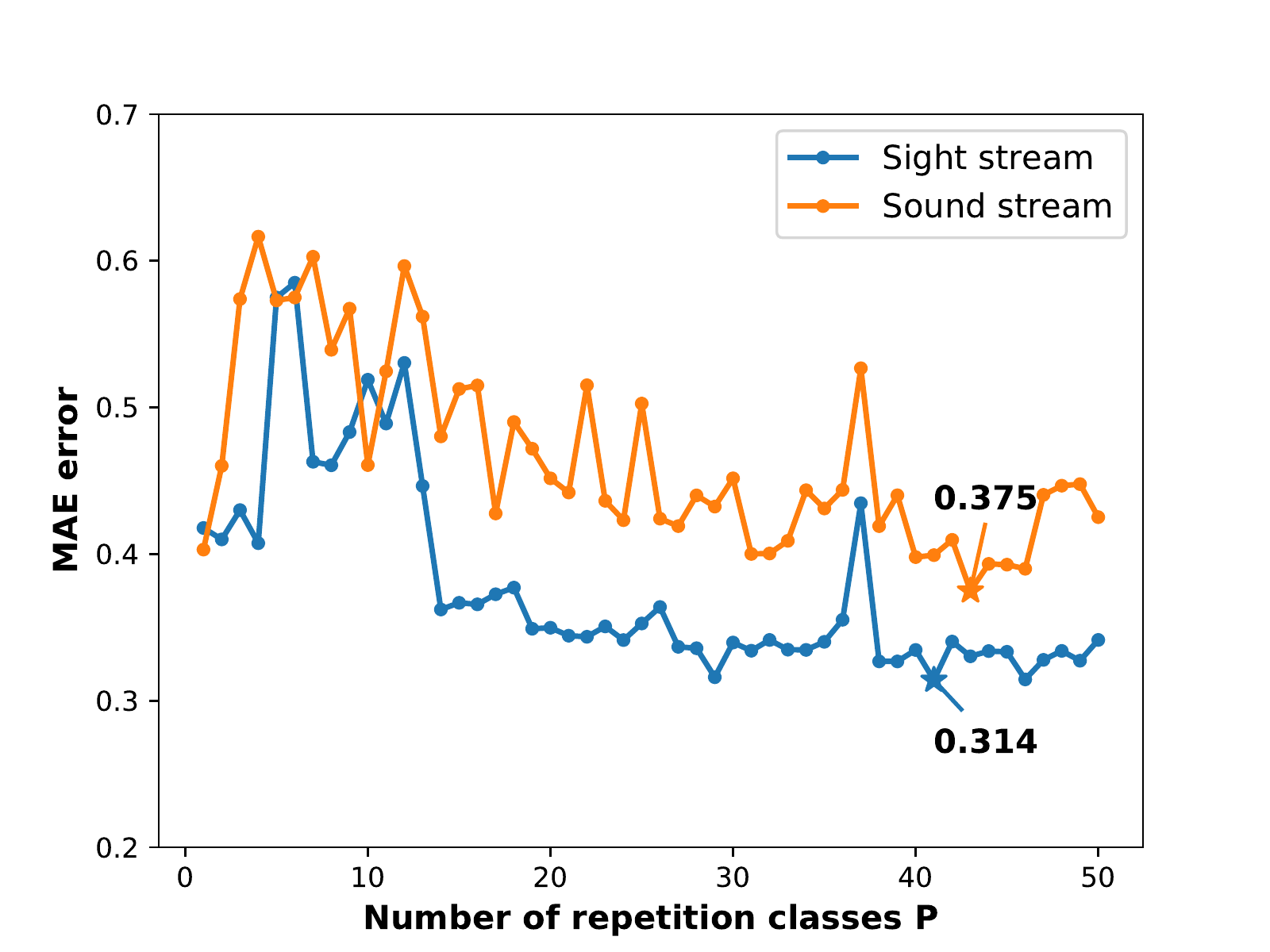}
\vspace{-1mm}
\caption{\textbf{Effect of repetition classes.} The performances of both streams fluctuates slightly when $P$ is large enough, and achieves the best result when $P{=}41$ (sight) and $P{=}43$ (sound). }
\label{fig:effectofp}
\end{figure}

\noindent \textbf{Effect of repetition classes.} 
As detailed in Eq.~\ref{eq:count1} and \ref{eq:count2}, our counting models for both modalities involve a parameter $P$, \ie the number of repetition classes. 
We evaluate its effect on both the sight and sound models. To this end, we fix the backbone architectures and train them by setting $P$ from $1$ to $50$ with the same setting described in Section~\ref{sec:implementation_details}. 
Here, we do not include cross-modal interactions, \ie the sight and sound models are trained and evaluated on Countix and our Countix-AV datasets, separately. 
The results are shown in Figure~\ref{fig:effectofp}. 
The performances of both models are inferior when $P$ has a low value, demonstrating the need to model various types of repetitions. The performance fluctuates only slightly when $P$ is between 20 and 50. 
We observe that there is a spike when $P{=}36$ for both streams, at where the networks converge to local minimum and this can be eliminated by more training epochs. 
In particular, the sight and sound models obtain their best results at $P{=}41$ and $P{=}43$, the values we use for all other experiments, while Countix~\cite{zisserman2020} and our Countix-AV cover $45$ and $19$ action categories. 
Thus, the repetition types do not simply correspond to the number of action categories. For instance, in Figure~\ref{fig:qualitative2}, the second and the last rows have similar repetition classification distributions, while dissimilar to the fourth row. We show more examples in the supplementary material.

\begin{table*}[!t]
\centering
\resizebox{0.92\linewidth}{!}{
\begin{threeparttable}
    \begin{tabular}{lccccccc}
    \toprule
     & \multicolumn{2}{c}{\textbf{UCFRep}} & \multicolumn{2}{c}{\textbf{Countix}}  & \multicolumn{2}{c}{\textbf{Countix-AV}} & \multicolumn{1}{c}{\textbf{Extreme Countix-AV}}\\
      \cmidrule(lr){2-3} \cmidrule(lr){4-5} \cmidrule(lr){6-7} \cmidrule(lr){8-8}
          & \textit{MAE~$\downarrow$} &  \textit{OBO~$\uparrow$} & \textit{MAE~$\downarrow$} &  \textit{OBO~$\uparrow$} & \textit{MAE~$\downarrow$} &  \textit{OBO~$\uparrow$} & \textit{MAE~$\downarrow$} \\
        \midrule
                Baseline$^{\dagger}$ & 0.474 & 0.371 & 0.525 & 0.289 & 0.503 & 0.269 & 0.620 \\
                Dwibedi \etal~\cite{zisserman2020} &- &- & 0.364 & \textbf{0.697} & -& -& -\\
                Levy and Wolf*~\cite{Levy2015} & 0.286  & 0.680 &- & - & -& -& -\\
                Zhang~\etal~\cite{context2020} & 0.147  & 0.790 & - & - & - & - & - \\ 
                \textit{This paper:} \textbf{\textit{Sight}} & \textbf{0.143 } & \textbf{0.800} & 0.314 & 0.498  & 0.331 & 0.431 & 0.392 \\
                \textit{This paper:} \textbf{\textit{Sound}} & - & - & 0.793 & 0.331  & 0.375 & 0.377 & 0.351 \\
                \textit{This paper:} \textbf{\textit{Sight \& Sound}} & - & - & \textbf{0.307} & 0.511 & \textbf{0.291} & \textbf{0.479} & \textbf{0.329}\\

    \bottomrule
    \end{tabular}
           \footnotesize{$^{\dagger}$~Sight-only model, pre-trained on Countix, publicly released by authors of~\cite{zisserman2020}.}\\
        \footnotesize{*~Sight-only model~\cite{Levy2015}, reproduced and pre-trained on UCFRep by authors of~\cite{context2020}.} \\
    \end{threeparttable}
    }
\renewcommand\thetable{5}
\caption{\textbf{Comparison with state-of-the-art.} Our sight model outperforms two recent state-of-the-art repetition counting algorithms in terms of MAE, while combining sight and sound shows further benefit in reducing counting error, especially in visually challenging settings. }
\label{tab:comparison_v}
\vspace{-4.5mm}
\end{table*}

\begin{table}[t!]
\centering
\resizebox{\linewidth}{!}{
\begin{tabular}{lccc}
\toprule
\textbf{Vision challenge} & \textbf{Sight}  & \textbf{Sound} & \textbf{Sight \& Sound} \\
\midrule
Camera viewpoint changes & 0.384 & 0.376 & 0.331 \\
Cluttered background & 0.342 & 0.337 & 0.307 \\
Low illumination & 0.325 & 0.269 & 0.310 \\
Fast motion & 0.528 & 0.311 & 0.383 \\
Disappearing activity & 0.413 & 0.373 & 0.339 \\
Scale variation  & 0.332 & 0.386 & 0.308 \\
Low resolution  & 0.348 & 0.303 & 0.294 \\
\midrule
\textit{Overall} & 0.392 & 0.351 & 0.329 \\
\bottomrule
\end{tabular}
}
\renewcommand\thetable{4}
\caption{\textbf{MAE metric on hard cases in repetition counting}. Sound tends to have lower MAE than sight. Combining sight and sound always outperforms sight only. }
\label{tab:results_Extreme}
\vspace{-5.5mm}
\end{table}

\noindent \textbf{Effectiveness of temporal stride module. }
We also considered fixed temporal strides for the sight-stream on Countix. MAE varies from 0.607 to 0.378 (see appendix). Our temporal stride decision module obtains a better 0.314 MAE.

\noindent \textbf{Hard cases in repetition counting.} 
To quantitatively evaluate the contribution of sound information and how sensitive the sight stream is under different visually challenging environment, we test the sight, sound and full sight and sound model separately on the Extreme Countix-AV dataset. The results are listed in Table~\ref{tab:results_Extreme}. 
Compared to the performance on Countix-AV dataset, which is dominated by videos with normal sight conditions, the MAE of the sight stream increases considerably. In contrast, the sound stream performs stably and is superior under visually challenging circumstances as expected, except for the scale variation challenge. This means that changes in image quality can easily affect the sight stream. 
Especially when activities are moving fast or disappearing due to occlusions, the value of the sound stream is prevalent. 
Combining sight and sound is always better than sight only, resulting in considerable MAE reductions on videos with camera view changes, disappearing activities, scale variation and cluttered background. 
For scale variation, the sound stream does not perform competitively compared to the sight stream, while the fused results do improve over the sight stream. 
This again indicates the effectiveness of our reliability estimation module. 
For low illumination and fast motion, the sight stream performs poor compared to the sound stream, and the combination cannot improve over the sound stream only.  
Overall, the integration of sight and sound is better than unimodal models and more stable when the imaging quality varies.  %

\noindent \textbf{Comparison with state-of-the-art. }
We compare our method with two recent state-of-the-art (vision-only) repetition counting models~\cite{zisserman2020,context2020} and one early work by Levy and Wolf~\cite{Levy2015}, as shown in Table~\ref{tab:comparison_v}. As the complete code of~\cite{zisserman2020} is unavailable, we also report the performance of their released (vision-only) model as a baseline. 
Our sight-only stream already outperforms Dwibedi \etal~\cite{zisserman2020} on their original Countix dataset with respect to the MAE metric, %
and achieves competitive performance on UCFRep~\cite{context2020}. 
Note the work by Zhang~\etal~\cite{context2020} needs the training videos to have boundary annotations for each repetition, which are not provided for Countix~\cite{zisserman2020}. 
As Countix is dominated by ``silent" repetitions, our sound-only model performs inferior compared to the sight-only model. Nevertheless, our full sight and sound model sets a new state-of-the-art on all three Countix datasets in MAE and surpasses the released model of~\cite{zisserman2020} by a large margin. Therefore, we conclude that when the original sound track of the video is available, audiovisual repetition counting is superior to sight-only models. 

\section{Conclusion}
We propose to count repetitive activities in video by sight and sound using a novel audiovisual model. To facilitate further progress, we repurpose and reorganize an existing counting dataset for sight and sound analysis. %
Experiments show that sound can play a vital role, and combining both sight and sound with cross-modal temporal interaction is beneficial. Using sight only we already outperform the state-of-the-art in terms of MAE. When adding sound, results improve further, especially under harsh vision conditions. 

\newpage

{\small
\bibliographystyle{ieee_fullname}
\bibliography{egbib}
}
\clearpage

\appendix
\section*{Overview of Appendix}

\begin{table*}[h]
\centering
    \begin{tabular}{lrrrrrrrr}
    \toprule
       & \multicolumn{4}{c}{\textbf{Number of videos}} & \multicolumn{4}{c}{\textbf{Average count groundtruth}} \\
        \cmidrule(lr){2-5}   \cmidrule(lr){6-9}
          \textbf{Action class} & \textit{Train} &  \textit{Validation} & \textit{Test} & \textit{Total} &  \textit{Train} &  \textit{Validation} & \textit{Test} & \textit{Average}\\
        \midrule
                battle rope training  & 57 & 17 & 41 & 115 & 14 & 10 & 6 & 11 \\
                bouncing ball (not juggling) & 63 & 25 & 41 & 129 & 7 & 9 & 7 & 7 \\
                bouncing on trampoline & 22 & 7 & 15 & 44 & 6 & 6 & 5 & 6\\
                clapping & 16 & 14 & 37 & 67 & 7 & 7 & 9 & 8 \\
                gymnastics tumbling & 27 & 6 & 15 & 48 & 4 & 3 & 4 & 4 \\
                juggling soccer ball & 65 & 23 & 9 & 97 & 11 & 11 & 9 & 11\\
                jumping jacks & 31 & 16 & 24 & 71 & 7 & 5 & 5 & 6 \\
                mountain climber (exercise) & 37 & 12 & 22 & 71 & 10 & 9 & 10 & 10 \\
                planing wood & 37 & 16 & 25 & 78 & 5 & 6 & 5 & 5\\
                playing ping pong & 79 & 25 & 34 & 138 & 3 & 3 & 3 &3\\
                playing tennis & 42 & 11 & 24 & 77 & 3 & 3 & 3 & 3 \\
                running on treadmill & 51 & 13 & 24 &88 & 13 & 13 & 10 & 12 \\
                sawing wood & 55 & 12 & 41 & 108 & 9 & 7 & 7 & 8 \\
                skipping rope & 62 & 24 & 36 & 122 & 12 & 11 & 9 & 11 \\
                slicing onion & 110 & 40 & 66 & 216 & 12 & 13 & 11 & 12 \\
                swimming & 80 & 13 & 32 & 125 & 5 & 5 & 6 & 5\\
                tapping pen & 38 & 12 & 24 & 74 & 19 & 25 & 24 & 22\\
                using a wrench & 22 & 3 & 9 & 34 & 5 & 3 & 5 & 5\\
                using a sledge hammer & 93 & 22 & 43 &158 & 5 & 5 & 5 & 5\\
        \midrule
        Total & 987 & 311 & 562 & 1860 & - & - & - & -\\
    \bottomrule
    \end{tabular}
\renewcommand\thetable{6}
\caption{\textbf{Countix-AV dataset statistics.} Note our model does not use the action class labels.}
\label{tab:statistics_countixav}
\end{table*}

The inference procedures of our model are first illustrated in Appendix~\ref{sec:appendix_inference_procedures}. 
Then we present additional details of our Countix-AV and Extreme Countix-AV datasets in Appendix~\ref{sec:appendix_countixav} and Appendix~\ref{sec:appendix_extreme_countix_av}. 
The effect of hyperparameter $\theta_s$ and margin $m$ in our temporal stride decision module is evaluated in Appendix~\ref{sec:appendix_ablation}, as well as the comparison with fixed temporal strides. 
For the reliability estimation module, we also study the effect of two hyperparameters $\theta_r^v$ and $\theta_r^a$ and compare with two alternatives in Appendix~\ref{sec:appendix_reliability}. 
In Appendix~\ref{sec:appendix_class_sup}, we analyze the performance of training our model with action class supervision and we demonstrate the implementation details of our sight model on the UCFRep~[\textcolor{green}{42}] dataset in Appendix~\ref{sec:appendix_ucf_implementation}. 
We finally explain the videos provided in the supplementary material in Appendix~\ref{sec:appendix_repetition_classes} and Appendix~\ref{sec:appendix_example_videos}.

\section{{Inference Procedures}}

\label{sec:appendix_inference_procedures}
For each video, we first divide it into video clips with temporal strides of $\{1,2,...,S_K\}$ and their corresponding audio signals, which are sent into the networks simultaneously. In experiments, we find $S_K{=}5$ is enough for the used datasets, and it can be enlarged for situations where the action takes place slowly. Then, we choose the stride with the maximum score outputted by the temporal stride decision module to resample the video and preserve the estimated reliability score of the selected stride for later fusion. 
In the end, after obtaining the counting results from both streams, the final prediction of our model is computed by Eq.~\textcolor{red}{10}. 

\section{Countix-AV Dataset Statistics}

\label{sec:appendix_countixav}
The Countix-AV dataset is a subset of Countix~[\textcolor{green}{11}] and the videos come from YouTube. 
It includes a total number of 19 classes for which the repetitive actions have a clear sound. %
The statistics per class are summarized in Table~\ref{tab:statistics_countixav}, including the number of videos per train, val and test fold, as well as the average count ground truth per class and fold.  
Some example videos are included in the supplementary material (``Learned repetition classes/Sound").

\section{{Extreme Countix-AV Dataset Details}}

\label{sec:appendix_extreme_countix_av}
The Extreme Countix-AV contains 214 videos in total, with 156 from Countix-AV and 58 from the VGGSound dataset~[\textcolor{green}{8}]. 
We define 7 vision challenges to collect videos.  
First, we manually check every video and choose those that have camera viewpoint changes, disappearing activity and scale variation based on our visual observation. 
Then, for the cluttered background challenge, we also manually select the videos in which there are multiple persons appearing simultaneously while only one person is doing the repetitive actions or the object conducting repetitive activity is too small and hard to be distinguished (\eg, some videos of bouncing ball). 
To collect videos captured in low illumination, we compute the average pixel intensity of each video, and add those with values below 100 (in the range of 0 and 255) into our dataset. 
For the fast motion challenge, we compute the average period length of each video according to the counting annotations, and find the videos with the average period length shorter than 3 frames. 
Finally, together the videos of those 7 challenges form our Extreme Countix-AV dataset.

\begin{figure}[b!]
\centering
\includegraphics[height=0.5\linewidth,width=0.9\linewidth]{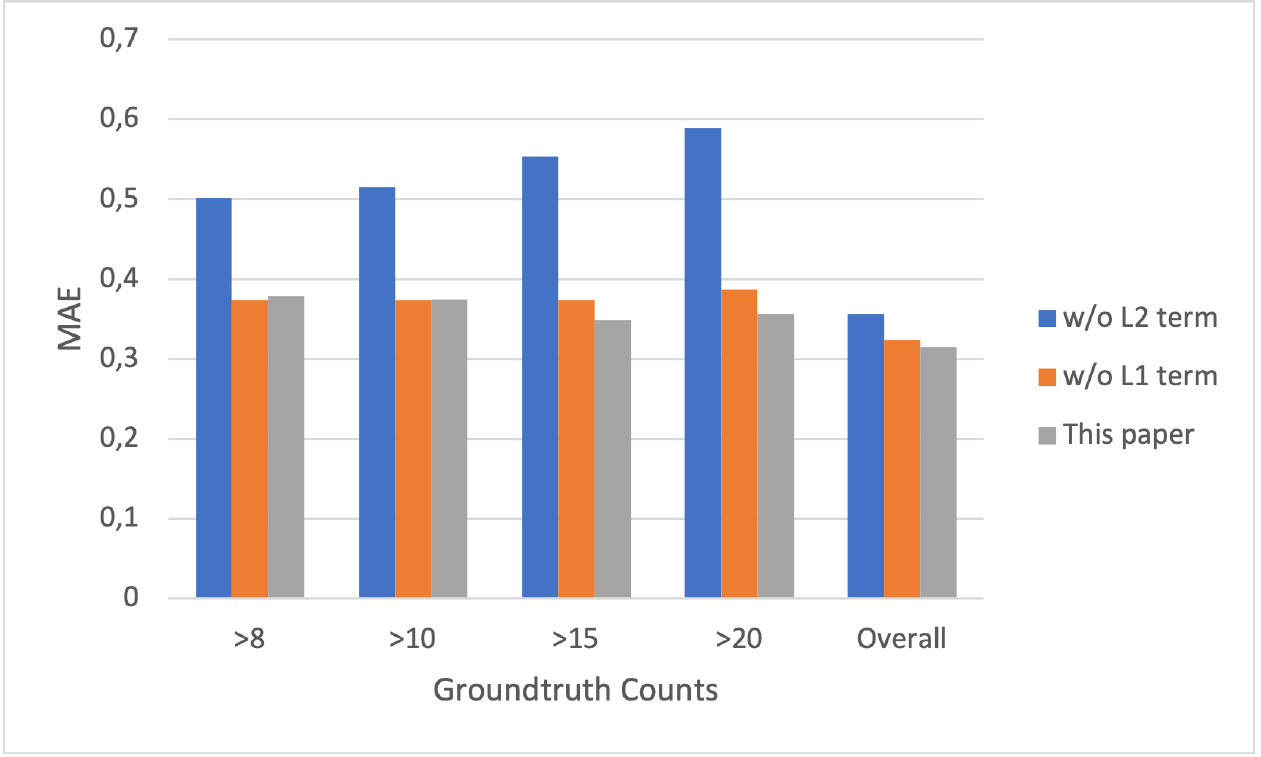}
\renewcommand\thefigure{5}
\caption{\textbf{Performance on videos with more repetitions.}}
\label{fig:appendix_more_repetitions}
\end{figure}

\begin{figure}[b!]
\centering
\includegraphics[height=0.5\linewidth,width=0.9\linewidth]{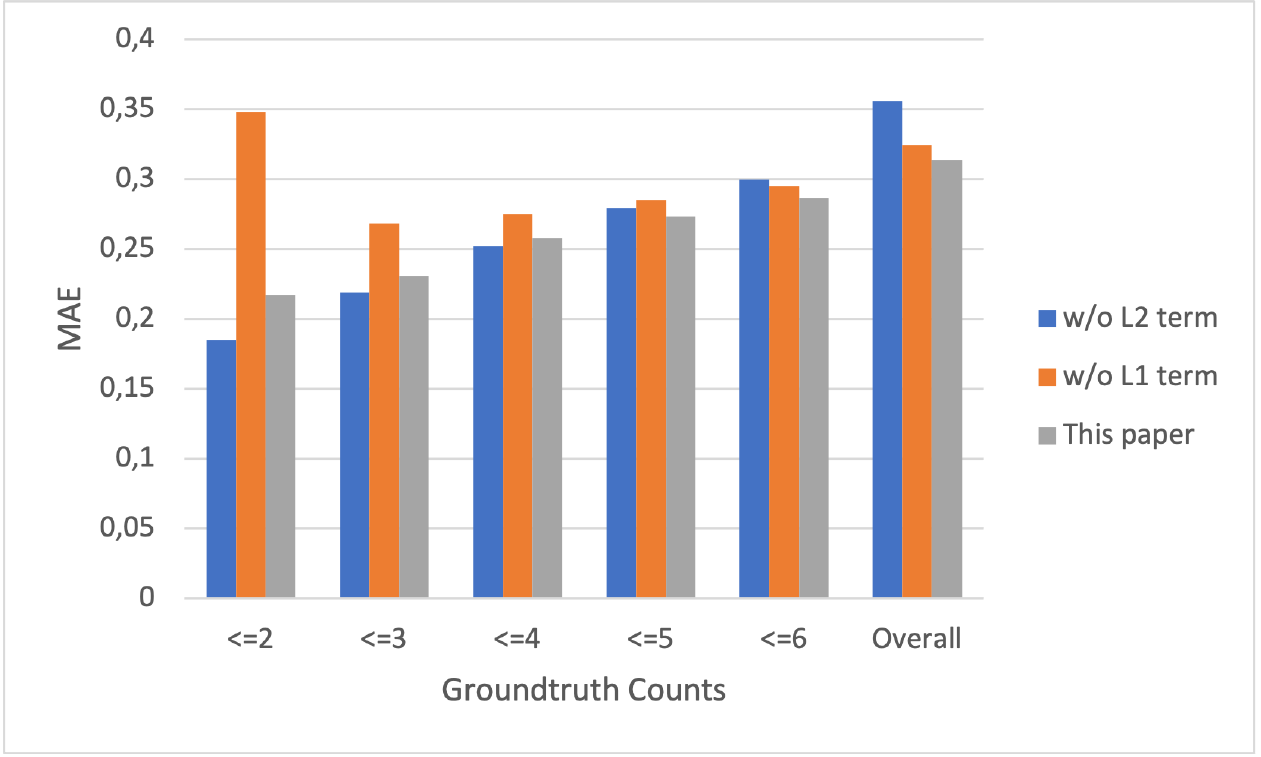}
\renewcommand\thefigure{6}
\caption{\textbf{Performance on videos with a few repetitions.}}
\label{fig:appendix_less_repetitions}
\end{figure}

\section{Effect of L1 and L2 Loss Terms}

In Eq.~\textcolor{red}{6} and Eq.~\textcolor{red}{7}, 
both L1 and L2 terms are used in the loss functions of the sight and the sound streams to balance accuracy on small and large counts. 
To illustrate their effectiveness, we perform an ablation on different term combinations and report the sight-only model performance on Countix~[\textcolor{green}{11}] towards videos of various groundtruth counts. 
As shown in Figure~\ref{fig:appendix_more_repetitions}, for videos with many repetitions, results do not degrade due to the L1 loss, with 0.356 MAE for videos with more than 20 cycles compared to 0.387 (w/o L1 loss) and 0.553 (w/o L2 loss). However, for videos with few repetitions, results get worse without L1 loss as shown in Figure~\ref{fig:appendix_less_repetitions}. The MAE for the sight stream increases from 0.217 to 0.348 on videos having only 2 repetitions. 
Therefore, the combination of L1 and L2 terms results in the best overall performance.

\section{Sight Stream Results}

\label{sec:appendix_ablation}

Here, we study the effect of hyperparameter $\theta_s$ and margin $m$ in our temporal stride decision module. 
All the experiments are based on the sight stream with visual modality only and the original Countix~[\textcolor{green}{11}] dataset. 
Note that despite the performance varies under different settings, all the results by our sight stream outperform the state-of-the-art by Dwibedi \etal~[\textcolor{green}{11}] considerably.

\begin{figure}[h!]
\centering
\includegraphics[height=0.7\linewidth,width=0.9\linewidth]{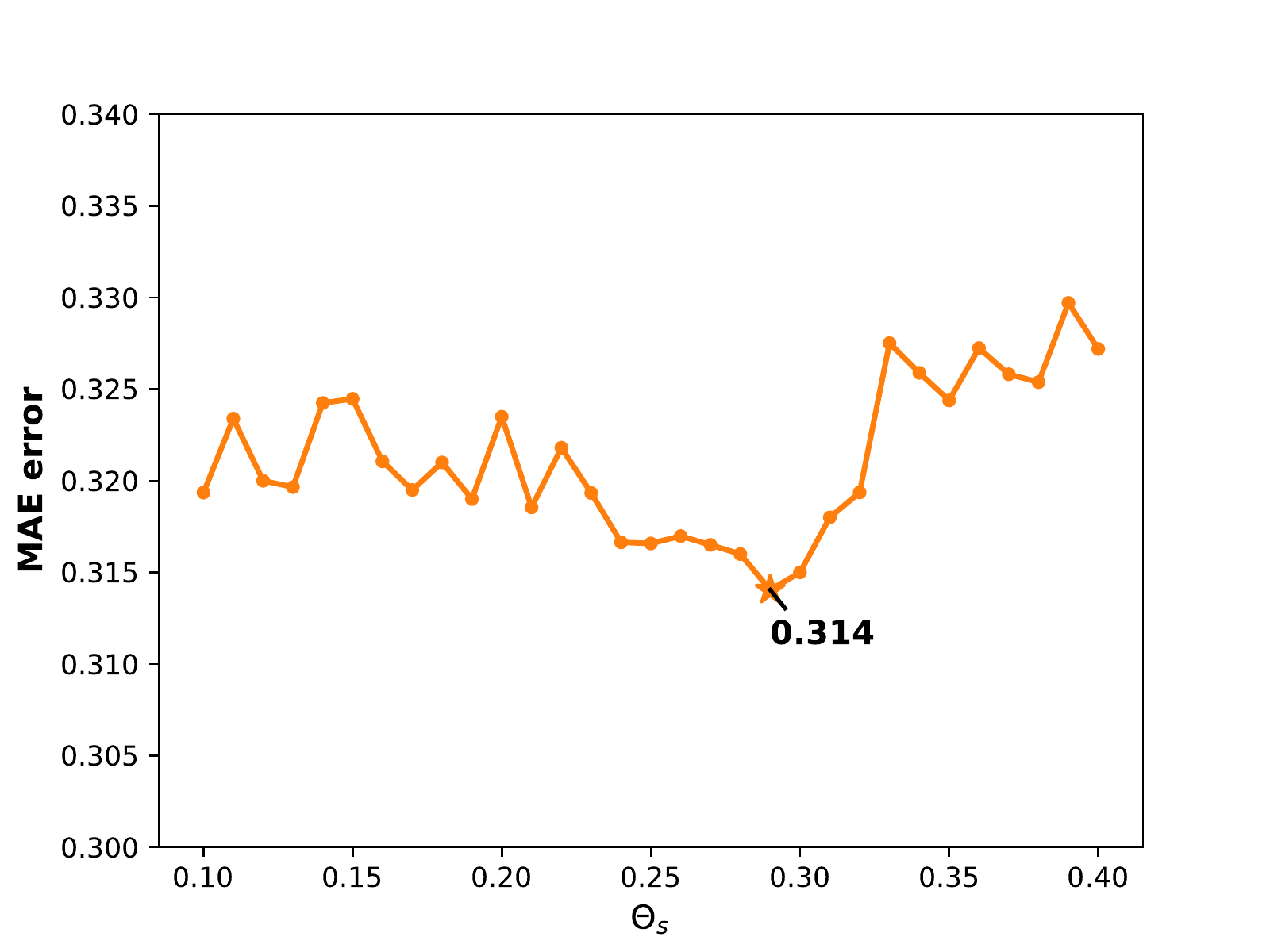}
\renewcommand\thefigure{7}
\caption{\textbf{Effect of $\theta_s$.} The performance of the sight stream is not very sensitive to $\theta_s$ and the best result is achieved at $\theta_s{=}0.29$. }
\label{fig:stride_theta_s}
\end{figure}

\paragraph{Effect of $\theta_s$. }
As illustrated in Section~\textcolor{red}{3.3}, 
$\theta_s$ is used to select the negative strides for training. 
With a higher $\theta_s$, the chosen negative strides are larger and lead the sight stream to have more omissions. 
In contrast, a small $\theta_s$ makes the selection rule strict and thus results in over-fit issues. 
We study the effect of $\theta_s$ by setting it in the range of [0.1, 0.33), and the results are shown in Figure~\ref{fig:stride_theta_s}. 
We can conclude that the performance is not very sensitive to $\theta_s$, and empirically $\theta_s{=}0.29$ represents the best trade-off. 
We also observe that the average MAE error increases when $\theta_s \geq 0.33$, since the trained sight stream tends to select larger temporal strides and omit certain repetitions.

\begin{figure}[b!]
\centering
\includegraphics[height=0.7\linewidth,width=0.9\linewidth]{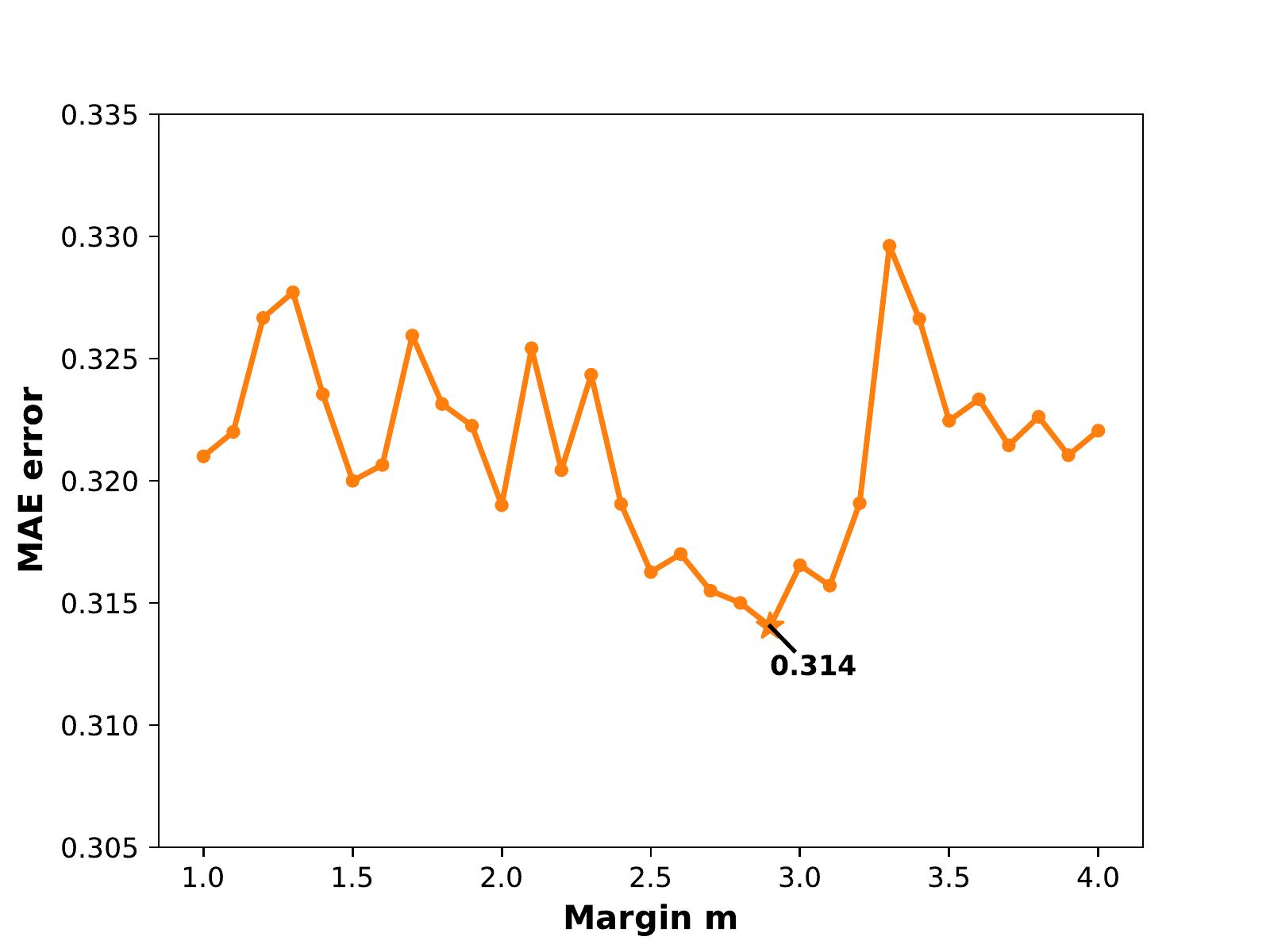}
\renewcommand\thefigure{8}
\caption{\textbf{Effect of margin $m$ for the max-margin ranking loss.} The value of $m$ does not influence the performance much and $m{=}2.9$ results in the lowest MAE error. }
\label{fig:stride_margin}
\end{figure}

\paragraph{Effect of margin $m$. }As detailed in Section~\textcolor{red}{3.3}, 
the max-margin ranking loss is adopted for training. In Figure~\ref{fig:stride_margin}, we show the performance of the sight stream when $m$ varies from 1.0 to 4.0. 
We can see that the MAE error fluctuates between 0.314 and 0.330, so the value of margin $m$ does not affect the results much.

\paragraph{Effectiveness of temporal stride module. }We report sight-stream results for fixed temporal strides on Countix in Table~\ref{tab:appendix_temporal_strides}. Our temporal stride decision module obtains a much better 0.314 MAE.

\begin{table}[t!]
\centering
\resizebox{\linewidth}{!}{
\begin{tabular}{ccccccc}
\toprule
\textbf{Temporal stride} & 1  & 2 & 3 & 4 & 5 & 6 \\
\midrule
\textit{MAE~$\downarrow$} & 0.607 & 0.378 & 0.387 & 0.427 & 0.467 & 0.475 \\
\bottomrule
\end{tabular}
}
\renewcommand\thetable{7}
\caption{\textbf{Effectiveness of temporal stride module. }Using a fixed temporal stride results in inferior performance compared to 0.314 MAE by our temporal stride module.}
\label{tab:appendix_temporal_strides}
\end{table}

\section{Reliability Estimation Module Results}
\label{sec:appendix_reliability}
\paragraph{Effect of $\theta_r^v$ and $\theta_r^a$. }
In Section~\textcolor{red}{3.4}, 
we use two thresholds $\theta_r^v$ and $\theta_r^a$ to collect predictions from both modalities for training the reliability estimation module. 
To study the effect of these two thresholds, we set them to different values and the results are shown in Table~\ref{tab:effectof2thresholds}. 
It is clear that the MAE fluctuates slightly between 0.291 and 0.296 under various settings, and the best performance is achieved when $\theta_r^v{=}0.360$ and $\theta_r^a{=}0.400$. 
In particular, the reliability estimation module is always superior to simply averaging predictions.

\begin{table}[t!]
\centering
\begin{tabular}{ccc}
\toprule
\textbf{$\theta_r^v$} & \textbf{$\theta_r^a$} & \textit{MAE~$\downarrow$}\\
\midrule
0.365 & 0.420 & 0.294\\
0.365 & 0.400 & 0.291 \\
0.365 & 0.390 & 0.292  \\
0.365 & 0.380 & 0.294  \\
0.360 & 0.420 & 0.292  \\
\textbf{0.360} & \textbf{0.400} & \textbf{0.291}  \\
0.360 & 0.390 & 0.292  \\
0.360 & 0.380 & 0.295  \\
0.355 & 0.420 & 0.294  \\
0.355 & 0.400 & 0.293  \\
0.355 & 0.390 & 0.293  \\
0.355 & 0.380 & 0.296  \\
0.350 & 0.420 & 0.292  \\
0.350 & 0.400 & 0.293  \\
0.350 & 0.390 & 0.292  \\
0.350 & 0.380 & 0.295  \\
\bottomrule
\end{tabular}
\renewcommand\thetable{8}
\caption{\textbf{Effect of thresholds $\theta_r^v$ and $\theta_r^a$ in the reliability estimation module.} The performance of the full sight and sound model varies slightly with different thresholds, but is always better than simply averaging the predictions from both streams. }
\label{tab:effectof2thresholds}
\end{table}

\begin{table}[h]
\centering
\begin{tabular}{lcc}
\toprule
\textbf{Model components} & \textit{MAE~$\downarrow$} & \textit{OBO~$\uparrow$}  \\
\midrule
Predictions from the final models & 0.297 & 0.436 \\
Fully connected layer & 0.301 & 0.421 \\
Full sight and sound model & 0.291 & 0.479 \\
\bottomrule
\end{tabular}
\renewcommand\thetable{9}
\caption{\textbf{Comparison with other fusion schemes} on Countix-AV. Using empirical predictions is better than the predictions from final models and our fusion scheme by predicting reliability score performs superior to two additional fully connected layers for feature integration and counting prediction. }
\label{tab:fusion_comparison}
\end{table}

\paragraph{Comparison with other fusion methods. }
To illustrate the superiority of our proposed scheme, which uses empirical predictions for training, we compare our approach with two alternatives. One is to directly use the predictions from the final trained models over the training videos for learning. The other is similar to the fusion method described in~[\textcolor{green}{7}] that trains two additional fully connected layers working in tandem built upon the penultimate layers of both sight and sound streams, which take the concatenated features from both modalities as inputs. 
As our original counting model, one fully connected layer outputs the repetition classification results and the other predicts the counting result of each class. The loss function is the same as Eq.~\textcolor{red}{6} and Eq.~\textcolor{red}{7} 
with the same hyperparameters but $P$ is set to $41$ for the best result. 

The results are shown in Table~\ref{tab:fusion_comparison}. We observe empirical predictions perform better than directly adopting predictions from the final models, and our reliability estimation module outputting the reliability score outperforms the counterpart that uses fully connected layers for feature fusion as well as counting prediction. Therefore, our proposed reliability estimation scheme effectively integrates information from both modalities for more accurate counting prediction.

\section{Counting with Action Class Supervision } 

\label{sec:appendix_class_sup}

To verify whether the action class labels can improve the counting accuracy further, we replace the $L_{div}$ in Eq.~\textcolor{red}{2} and Eq.~\textcolor{red}{3} 
with a cross-entropy loss using action class labels for supervision to train the repetition classification branch. 
The results are shown in Table~\ref{tab:known_classes}, 
while the performances of our original models can be found in Table~\textcolor{red}{5}. 
We observe that action class supervision can only improve the counting accuracy of the sight stream by a small margin, while degrade the performance of the sound stream and the full sight and sound model. 
The results demonstrate that repetition classes cannot be simply regarded as action classes. 
Although action class supervision can guide the network to count the correct repetitive movements inside each video, each action class may contain various repetition classes (\ie repetitive motions) in different videos, which should not be counted in the same way. For instance, for the sight stream, videos of doing aerobics contain different movements that needed to be counted. In contrast, the arm shows similar motions in some videos belong to action classes of playing table tennis and playing tennis. 
Similar phenomenon can also be found in the sound stream. 
On one hand, some videos of ``Slicing onion" and ``Tapping pen" contain similar sound patterns and thus can be counted in the same way. 
On the other hand, the sound stream needs to focus on various tones in different videos that belong to the action class ``skipping rope". 
In some videos, it is easy and reliable to count the repetitions by hearing how many times the feet of the person touch the ground. 
However, in some other videos, only the sound of rope is clear and usable. 
We provide example videos of learned repetition classes in the folder ``Learned Repetition Classes" of the supplement for both sight and sound streams with illustration in Appendix~\ref{sec:appendix_repetition_classes}. 
Therefore, we can conclude that for temporal repetition counting, our automatically learned repetition classification layer is superior to its counterpart that uses action class supervision.

\begin{table}[t!]
\centering
\resizebox{\linewidth}{!}{
    \begin{tabular}{lcccc}
    \toprule
       &\multicolumn{2}{c}{\textbf{Countix}} & \multicolumn{2}{c}{\textbf{ Countix-AV}} \\
      \cmidrule(lr){2-3} \cmidrule(lr){4-5}
          & \textit{MAE~$\downarrow$} &  \textit{OBO~$\uparrow$} & \textit{MAE~$\uparrow$} &  \textit{OBO~$\uparrow$} \\
        \midrule
                Sight  & 0.309 & 0.490 & 0.330 & 0.407\\
                Sound &- & - &0.400& 0.301\\
                Sight \& Sound & -  & -  & 0.316 & 0.424 \\ %
    \bottomrule
    \end{tabular}
}
\renewcommand\thetable{10}
\caption{\textbf{Counting with action class supervision.} Only the sight stream can benefit from the action class supervision marginally, while the performances of the sound stream and the full sight and sound model degrade. Therefore, action class supervision cannot effectively guide the learning of repetition counting models. }
\label{tab:known_classes}
\end{table}

\section{Implementation Details for UCFRep}
\label{sec:appendix_ucf_implementation}
Here, we illustrate the training details of our sight-only model on the UCFRep~[\textcolor{green}{42}] dataset. 
Similar to the training on the Countix~[\textcolor{green}{11}] dataset, we also initialize the weights of the model from a Kinetics pretrained checkpoint. 
Hyperparameters, like $\lambda_v^1$, $\lambda_v^2$, margin $m$, batch size, learning rate, etc, remain the same, as described in Section~\textcolor{red}{4.3}. 
The only difference is the number of repetition classes $P$, which is adjusted by greedy search, and we find $P{=}24$ works best. 

\section{Learned Repetition Classes}

\label{sec:appendix_repetition_classes}
As our model learns to classify the input videos into different repetition classes automatically during training, here we visualize these learned classes by example videos in the supplementary material. 
The sight and sound streams are illustrated separately. 

\paragraph{Learned repetition classes of the sight stream. }
 In the folder ``Learned Repetition Classes/Sight" of the supplementary material, we prepare 4 groups, named from ``1.mp4" to ``4.mp4", for the illustration of videos which have similar repetition class distributions from the repetition classification layer but belong to different action classes. 
Similar movements can be discovered in videos of each group despite there are significant variations in appearance. 
In ``Doing aerobics.mp4", we can see that the video segments contain different repetitive motion patterns and thus they are classified into different repetition classes for counting by our sight model. However, in the field of action recognition, these segments belong to the same action class ``doing aerobics".

\paragraph{Learned repetition classes of the sound stream. }
Similar to the sight stream, in the folder ``Learned Repetition Classes/Sound" of the supplementary material, we prepare 4 groups named from ``1.mp4" to ``4.mp4", in which videos of each group have similar class distributions by the repetition classification layer. 
It is clear that the audio tracks in each group have similar sound patterns but belong to various action classes. 
We also present some videos belong to the the same action class (\ie skipping rope) but are treated as different repetition classes due to various types of sound in ``skipping\_rope.mp4".

\section{Example Videos}

\label{sec:appendix_example_videos}

In ``demo\_video.mp4" of the supplementary material, we show some example videos of our Extreme Countix-AV dataset with corresponding challenges as well as the predictions from the sight and the sound stream, our full sight and sound model and groundtruth.

\end{document}